\documentstyle[twoside]{article}
\begin{document}
\def\Vox{$^{\fbox {\/}}$}
\def\prom{
\newtheorem{prop}{Proposition}[section]
\newtheorem{cor}[prop]{Corollary}
\newtheorem{lem}[prop]{Lemma}
\newtheorem{exa}[prop]{Example}
\newtheorem{sch}[prop]{Scholium}
\newtheorem{rem}[prop]{Remark}
\newtheorem{axi}[prop]{Axiom System}
\newtheorem{con}[prop]{Conjecture}
\newtheorem{theor}[prop]{Theorem}
\newtheorem{cla}[prop]{Claim}
\newtheorem{tef}[prop]{Definition}}
\def\cosdo{
\author {\ \\N.\ C.\ A.\ da Costa\\F.\ A.\ Doria\\\ \\Research Group
on Logic and Foundations,\\Institute for Advanced Studies, University of S\~ao
Paulo.\\Av.\ Prof.\ Luciano Gualberto, trav.\ J, 374.\\05655--010 S\~ao Paulo SP
Brazil.\\{\sc ncacosta@usp.br}\\\ \\Research Center on Mathematical Theories of
Communication\\and {\it Project Griffo},\\School of Communications,\\Federal
University at Rio de Janeiro.\\Av.\ Pasteur, 250. 22295--900 Rio RJ
Brazil.\\{\sc doria@omega.lncc.br}}}
\prom
\newcommand{\forc}{\,|\!\vdash}
\title {On the existence of certain total recursive functions in nontrivial axiom systems, I.\thanks
{Partially supported by CNPq, by FAPESP and by the PREVI Program, Federal University at Juiz de
Fora.}}
\cosdo
\date {April 1998\\Version 9.5}
\maketitle

\begin {abstract}
\noindent We investigate the existence of a class of ZFC--provably total recursive unary functions,
given certain constraints, and apply some of those results to show that, for $\Sigma _1$--sound set
theory, ${\rm ZFC}\not\vdash P<NP$. 
\end {abstract}

\pagestyle {myheadings}

\thispagestyle {empty}
\newpage

\

\tableofcontents
\thispagestyle {empty}
\newpage
\thispagestyle {plain}

\

\vspace {.5in}
\markboth {da Costa, Doria}{Total recursive function, I}
\section {Introduction}
\markboth {da Costa, Doria}{Total recursive function, I}

It is a classical fact that there are total unary recursive functions in the standard model for
arithmetic which cannot be proved to be so within reasonably strong axiomatic systems (see \cite
{ML}, p.\ 257 and references therein). Such a phenomenon can especially be seen to happen within PA
(Peano arithmetic) or within ZFC (Zermelo--Fraenkel set theory with the axiom of choice) if they
satisfy a soundness condition, $\Sigma _1$--soundness (see Definition \ref {sig}). 

We consider in this paper a more particular situation: what can be said about a given partial
recursive function when we add several specific restrictions to the objects considered within those
$\Sigma _1$--sound systems? We obtain a partial unary recursive function which, if it is true that
such a function is total, then that fact cannot be proved within those axiom systems, supposed
consistent. 

The original motivation for this paper stems out of the authors' search for incompleteness results
with respect to nontrivial ``rich'' axiom systems, such as ZFC or a sizable fragment of it, which
arise out of expressions for mathematical data, that is, which depend on the way we linguistically
encode the mathematical objects. Some rather striking undecidable statements may arise in that way,
and they settle noted questions as the integrability problem in classical mechanics, the decision
problem for chaotic systems \cite {CosDo1}, or Arnol'd's problems on the nature of stability in
dynamical systems \cite {CosDo5,CosDo10}. Those results not only  settle open questions, but do so
in a linguistic way, that is, they depend on the explicit form of expressions for mathematical
concepts that may arise within a formal system. Moreover they are technically simple like the
results presented here, and independent of the actual precise formal environment where they are
placed. 

We develop here the first part of a different (but also linguistically conceived) technique for the
construction of independence results. The present technique comes down from a standard diagonal
argument (see \cite {Ku}, p.\ 54, Ex.\ 5), and deals with provably recursive, denumerable objects
which are kept so throughout the procedure. That technique is again rather insensitive to the actual
formal setting where its proofs are supposed to happen, as we only require constructs from formal
arithmetic in it.  

We ask for the existence of a certain set of total recursive functions whose domain is informally
understood as a set of expressions that describe Turing machines. We show here that it is consistent
(with ZFC, taken as the formal background and supposed $\Sigma _1$--sound) to assume that those
functions do not exist within the required axiom system. 

As an application we discuss in the last Section the characterization problem for nondeterministic
polynomial computations. We show how that question relates to the present work, as well as to
previous work in the area. We then show that, if ZFC is $\Sigma _1$--sound, then
${\rm ZFC}\not\vdash\,P<NP$. Related and complementary results will be considered in Part II of this
paper. 

\subsection*{Conventions, notations, Turing machines}
\addcontentsline {toc}{subsection}{Conventions, notations, Turing machines}

\begin {rem}\rm 
The formal setting is ZFC (Zermelo--Fraenkel set theory with the axiom of choice); informal
definitions and proofs happen as usual within ordinary mathematics. Here we use Kleene's $T$ 
predicate \cite {IM,ML}; $\omega$ is the set of natural numbers. 

Restricted bounded arithmetic variables are in general used without making explicit their domains,
since they are always clear by context. \Vox
\end {rem}

\begin {rem}\label {TM}\rm 
We require Turing machines in our argument; their behavior is specified below to avoid ambiguities:
\begin {enumerate}
\item Our Turing machines are defined over the set $A_2^*$ of finite words on the binary alphabet
$A_2=\{0,1\}$. 
\item Each machine has $n+1$ states $s_0, s_1,\ldots, s_n$, where $s_0$ is the final state. (The
machine stops when it moves to $s_0$.)
\item The machine's head roams over a two--sided infinite tape. 
\item Machines input a single binary word and either never stop or stop when the tape has a
finite, and possibly empty set of binary words on it.  
\item The machine's {\it output word} will be the one over which the head rests when $s_0$ is
reached. (If the head lies on a blank square, then we take the output word to be the empty word.) 
\end {enumerate}

It is easy to formalize the concept of Turing machine within ZFC, as a Turing machine is a
mathematical structure. For examples of such formalizations see \cite {CosDo1}. In what
follows we suppose that we are given the canonical enumeration of binary strings
$\emptyset,0,1,00,01,10,11,\ldots$, and that each binary string is coded by the corresponding numeral
in the sequence $0,1,2,3,\ldots$. \Vox
\end {rem}

\markboth {da Costa, Doria}{Total recursive function, I}
\section{Preliminary developments}
\markboth {da Costa, Doria}{Total recursive function, I}

\begin {tef}\label {sig}
A recursively axiomatizable theory $R$ which is sufficient to develop elementary number theory is
$\Sigma _1$--{\bf sound} if and only if whenever $P(x)$ is a primitive recursive predicate such
that $R\vdash\exists x\,P(x)$, there is a natural number $n$ such that $P(n)$ holds. \Vox
\end {tef}

\begin {rem}\rm We will make explicit when we suppose that either PA or ZFC are $\Sigma _1$--sound.
\Vox 
\end {rem} 

\begin {tef}\label {bu}\ 

\begin {enumerate}
\item A {\rm ZFC} unary function $f$ is {\bf {\rm ZFC}--provably total recursive} if for some
G\"odel number $e_f$ for $f$, ${\rm ZFC}\vdash\forall x\in\omega\,\exists
z\in\omega\,(T(e_f,x,z)\wedge\,\forall y\in\omega\, f(y)=\{e_f\}(y))$  {\rm \cite {Sol}}. 
\item A {\rm ZFC}--unary predicate $Q(x)$ is {\bf {\rm ZFC}--provably total
recursive} if its characteristic function is {\rm ZFC}--provably total recursive. Extension of that
definition to a $n$--place predicate $Q(x_0,x_1,\ldots,x_n)$ is immediate. \Vox 
\end {enumerate}\end {tef}

We need a set ${\cal G}$ of ZFC--provably total recursive unary (t.r.u.) functions:

\begin {tef}\label {G}
The functions $g_{n}\in{\cal G}$, where $n\in\omega$, satisfy, for any $x$ and any $m , n$:
\begin {enumerate}
\item $g_{n}$ is {\rm ZFC}--provably t.r.u. 
\item $g_0(x) >2$.
\item $g_{n}(x+1)>g_{n}(x)$. 
\item $g_{n}(x)>g_{m}(x)$, $n > m$. \Vox
\end {enumerate}
\end {tef}

It is easy to see that infinitely many such sets ${\cal G}$ exist. 

Let ${\sf M}_0$, ${\sf M}_1$, ${\sf M}_2$, \ldots, be the single--tape Turing machines described
above, and ordered by their G\"odel numbers; we suppose that every $n\in\omega$ is a G\"odel number
for some machine.   

Let $[g_n(|x|)]$, $n\in\omega$, $|x|$ the length of $x$, an input to ${\sf M}_i$,
be a clock that stops ${\sf M}_i(x)$ over input $x$ after $g_n(|x|)-1$ cycles. We suppose that the
clock acts as follows: when the bound in the number of processing steps is reached, the clock stops
the machine to which it is coupled and then moves its state to $s_0$. The machine's output is the
word on which the head rests when the clock stops it. Let $\langle\ldots ,\ldots\rangle$ denote the
usual recursive 1--1 pairing function $\langle\ldots
,\ldots\rangle:\omega\times\omega\rightarrow\omega$. We define: 

\begin {tef}\label {X}
For $p=\langle i,n\rangle$, ${\sf P}_{p}=\langle {\sf M}_i,[g_n]\rangle$. ${\cal P}^0=\{{\sf P}_0,
{\sf P}_1, \ldots\}$; we order ${\cal P}^0$ by the code sequence $p=0,1,2,\ldots$ . \Vox
\end {tef}

\begin {rem}\rm 
${\cal P}^0$ contains all possible ordered pairs as in the preceding definitions; the number $p$
can intuitively be seen as coding an expression for a Turing machine. From here onwards we will
refer to those pairs when we talk about ``expressions for Turing machines'' of the kind considered
in the present paper. \Vox
\end {rem} 

\

\begin {tef}\ 

\begin {enumerate}
\item If there is a fixed $n\in\omega$ such that for every $x\in\omega$, ${\sf M}_i(x)$ stops before,
or at most at $g_n(|x|)$ machine cycles, we say that ${\sf M}_i$ is  ${\cal G}$--{\bf bounded}.
\item A total recursive unary function $f(x)$ is ${\cal G}$--{\bf bounded} if there is a ${\cal
G}$--bounded Turing machine that computes it. \Vox
\end {enumerate}
\end {tef}

\begin {prop}
If ${\sf M}_i$ is a ${\cal G}$--bounded machine, then there will be infinitely many ${\sf P}_p\in
{\cal P}^0$ which compute the same function as ${\sf M}_i$. \Vox
\end {prop}

\begin {rem}\rm 
We can of course allow index $n$ in $g_n$ to range over an initial segment of the ordinals, given a
notation system that satisfies the conditions spelled out in Definition \ref {G}. However we won't
require this fact in the present paper. \Vox
\end {rem}

\subsection*{The machine ${\sf V}$}
\addcontentsline {toc}{subsection}{The machine ${\sf V}$}

\begin {tef}\label {v} ${\sf V}(\langle x,s\rangle)$ is a fixed but otherwise arbitrary ${\cal
G}$--bounded Turing machine such that: 

\begin {enumerate}
\item ${\sf V}$ is an algorithm for a function $f_{\sf V}:\omega\times\omega\rightarrow\{0,1\}$. 
\item For every $x\in\omega$ there is an $s\in\omega$ such that ${\sf V}(\langle x,s\rangle)=1$.
Given the machine ${\sf P}_n$, whenever ${\sf P}_n(x)=s$ and ${\sf V}(\langle x,s\rangle)=1$, we
say that $x$ is {\bf acceptable for} $n$. 
\item For every $x\in\omega$ there is an $s\in\omega$ such that ${\sf V}(\langle x,s\rangle)=0$. 
\item For the empty word $\emptyset$ and for any nonempty strings $x$, $s$:
\begin {enumerate}
\item ${\sf V}(\langle\emptyset, s\rangle)=1$.
\item ${\sf V}(\langle x,\emptyset\rangle)=0$. 
\item ${\sf V}(\langle\emptyset,\emptyset\rangle)=0$. \Vox
\end {enumerate}
\end {enumerate}
\end {tef}

\begin {prop}\label {o}
Given a constant $x_0\in\omega$, there are infinitely many ${\sf P}_m$ such that $x_0$ is
acceptable for ${\sf P}_m$. 
\end {prop} 

{\it Proof}\,: Let $s$ be such that  ${\sf V}(\langle x_0,s\rangle)=1$. Let ${\sf C}_s$ be one of the
constant Turing machines that input anything and output $s$. Let $c_s(|x_0|)$ be its operation time
when input $x_0$ is fed to the machine. Then, given the conditions in Definition
\ref {G}, it is easy to see that $x_0$ is acceptable for the ${\cal G}$--bounded machine $\langle
{\sf C}_s, g_{c_s(|x_0|)}\rangle$. And consequently, it is acceptable for all similar pairs, with
clocks $[g_{(c_s(|x_0|) + k)}]$, for any natural $k$. 

Moreover, since we have agreed that for the empty word $\emptyset$, we always have 
${\sf V}(\langle x,\emptyset\rangle)=0$, any $x$, one immediately sees that there are infinitely
many ${\cal G}$--bounded Turing machines that, for a fixed $x_0$, will only accept that $x_0$, and no
other input. \Vox 

\subsection*{The predicate $P^{\phi}(m,x)$ and the function $f^{\phi}_P(m)$}
\addcontentsline {toc}{subsection}{The predicate $P^{\phi}(m,x)$ and the function $f^{\phi}_P(m)$}

In what follows we use the coding of binary sequences by natural numbers. We will now slightly
strengthen our notational conventions: 

\begin {tef}\label {fel"}
A {\bf representation} for ${\cal P}^0$ is any map $\phi :{\cal P}^0\rightarrow{\cal P}^0$,
such that:
\begin {enumerate}
\item $p\in\omega\mapsto\phi(p)\in\omega$ is a recursive permutation. 
\item ${\cal P}^0=\{{\sf P}^0_0,{\sf P}^0_1,{\sf P}^0_2,\ldots\}$, where ${\sf P}^0_0={\sf
P}_0, {\sf P}^0_1={\sf P}_1$, \ldots 
\item ${\sf P}^{\phi}_{\phi (p)}=_{\rm Def}\phi({\sf P}^0_p)$. 
\item $\phi{\cal P}^0 = {\cal P}^{\phi}=\{{\sf P}^{\phi}_0,{\sf P}^{\phi}_1,\ldots\}$. \Vox
\end {enumerate}
\end {tef}

\begin {tef}
For ${\sf P}^{\phi}_m\in{\cal P}^{\phi}$, 
$P^{\phi}(m,x)\leftrightarrow_{\rm Def}{\sf V}(\langle x,{\sf P}^{\phi}_m(x)\rangle)=0$. \Vox
\end {tef}

\begin {cor}
$P^{\phi}(m,x)$ is a {\rm ZFC}--provably recursive predicate. \Vox
\end {cor}

\begin {tef}\label {ph} $f^{\phi}_P (m)=_{\rm Def}\mu _x P^{\phi}(m,x)$. \Vox
\end {tef}

\begin {cor}
$f^{\phi}_{P}$ is partial recursive. \Vox
\end {cor}

\begin {rem}\rm
If ZFC is $\Sigma _1$--sound, then $f^{\phi}_P$ is also ZFC--provably partial recursive, as out of
the ZFC--provably recursive characteristic function for $P^{\phi}$ we can compute a G\"odel number
for a machine that computes $f^{\phi}_P$. \Vox
\end {rem}

\noindent One final result: 

\begin {cor}
${\rm ZFC}\vdash \forall m\in\omega\,\exists y\in\omega\,P^{\phi}(m,y)\leftrightarrow
f^{\phi}_P\mbox{\rm\ is total}$. \Vox
\end {cor}

\subsection*{Equivalence of representations}
\addcontentsline {toc}{subsection}{Equivalence of representations}

It doesn't matter which representation $\phi$ we choose:

\begin {prop}\label {ekiv}
${\rm ZFC}\vdash\forall m\,\exists x\, P^{\phi}(m,x)\leftrightarrow \forall n\,\exists x\, 
P^{\psi}(n,x)$. 
\end {prop}

{\it Proof}\,: $P^{\phi}(m,x)\leftrightarrow _{\rm Def}{\sf V}(x,{\sf P}^{\phi}_m(x))=0$. (We 
omit the $\langle\ldots\rangle$ for simplicity.) But ${\sf P}^{\phi}_m(x)={\sf P}^0_{\phi
^{-1}(m)}(x)$. (See Definition \ref {fel"}.)  Then:
$${\sf V}(x,{\sf P}^{\phi}_m(x))={\sf V}(x,{\sf P}^0_{\phi ^{-1}(m)}(x))=0.$$
Thus $${\rm ZFC}\vdash P^{\phi}(m,x)\leftrightarrow P^0(\phi^{-1}(m),x),$$ and ${\rm
ZFC}\vdash\exists x\,P^{\phi}(m,x)\leftrightarrow\exists x\,P^0(\phi^{-1}(m),x)$.  

To deal with $\forall$ recall that $\phi$ is 1--1 and onto. Then ${\rm
ZFC}\vdash\forall m\exists x\, P^{\phi}(m,x)\leftrightarrow\forall m\exists x P^0(\phi^{-1}(m),x)$,
and $${\rm ZFC}\vdash\forall m\,\exists x\, P^{\phi}(m,x)\leftrightarrow\forall
n\,\exists x\,P^0(n,x)$$ (with $n=\phi ^{-1}(m)$), whereas the conclusion, since the
argument holds for arbitrary $\phi$. \Vox 

\begin {cor}\label {ek2}
${\rm ZFC}\vdash\forall\phi\,[\phi\mbox{\rm\ is a representation}\rightarrow (\forall
m\,\exists y\,P^{\phi}(m,y)$ $\leftrightarrow f^{\phi}_P$ {\rm is total}$)]$. \Vox
\end {cor}

\markboth {da Costa, Doria}{Total recursive function, I}
\section {Nonexistence of a {\rm ZFC}--provably total $f^{0}_P$}
\markboth {da Costa, Doria}{Total recursive function, I}

In what follows we will frequently refer to ``machines'' for the sake of a more intuitive argument. 

\begin {prop}\label {sel}
There is a representation $\phi$ such that, if {\rm ZFC} is $\Sigma _1$--sound, for every
{\rm ZFC}--prov\-ab\-ly total recursive unary (t.r.u.) function $F_i$, ${\rm
ZFC}\vdash f^{\phi}_P\neq F_i$. 
\end {prop}

\begin {rem}\rm
Therefore either $f^{\phi}_P$ is partial or it is total but not ZFC--provably total. 

The idea of the proof is the following: we will recursively reorder the sequence ${\cal P}^0$ to
obtain ${\cal P}^{\phi}$ which is such that every ZFC--provably t.r.u.\ function will differ
somewhere from $f^{\phi}_P$ (see Definition \ref {ph}). More precisely: 
\begin {itemize}
\item Recall that $P^{\phi}(m,x)\leftrightarrow {\sf V}(\langle x,{\sf P}^{\phi}_m(x)\rangle)=0$
and that $f^{\phi}_P(m)=\mu _x P^{\phi}(m,x)$, where $m$ is the machine code in ${\cal P}^{\phi}$
and $x$ is the machine's input. 
\item If $y=\mu _x P^{\phi}(m,x)=f^{\phi}_P(m)$, then ${\sf V}(\langle y,{\sf
P}^{\phi}_m(y)\rangle)=0$, or equivalently by definition, $P^{\phi}(m,y)$ holds. 
\item Then if $y'=F_i(m)$, and if $\neg P^{\phi}(m,y')$  is provable, then that last statement
is equivalent to ${\sf V}(\langle y',{\sf P}^{\phi}_m(y')\rangle)=1$. 
\item Therefore $y'\neq y = f^{\phi}_P(m)=\mu _x P^{\phi}(m,x)$. 
\item Therefore $y'=F_i(m)\neq f^{\phi}_P(m)$.
\end {itemize}
We must arrange ${\cal P}^{\phi}$ so that for any $i$ and for at least one $m$, $y'\neq y$. \Vox
\end {rem}

{\it Proof}\,: We will consider two sequences of functions and machines: the first one, kept fixed
throughout the proof, is the sequence $F_0,F_1,\ldots$, of ZFC--provably t.r.u.\ functions. The
second sequence is the sequence ${\sf P}^0_0,{\sf P}^0_1,\ldots$ of ${\cal G}$--machines, which will
eventually be moved to and fro during the proof. 

Start theorem--proving in ZFC. Pick up all theorems of the form
\begin {equation}\label {equa}
{\rm ZFC}\vdash\forall x\in\omega\,\exists z\in\omega\,T(e,x,z)\,\wedge\,\forall
y\in\omega\,(f(y)=\{e\}(y)).
\end {equation}
(See Lemma \ref {Kr}.) We thus generate a r.e.\ sequence of G\"odel numbers $$e_{f_0}=e_0,
e_{f_1}=e_1, e_{f_2}=e_2,\ldots.$$ Repetitions are allowed---just pick up the G\"odel numbers as
they come up in the sequence. Out of them obtain the sequence of Turing machines
$${\sf M}_{e_0},{\sf M}_{e_1},\ldots$$ which compute the ${\rm ZFC}$--provably recursive total
functions $$\{e_0\}(x),\{e_1\}(x),\ldots$$ We use the notation: $F_j$ is computed by ${\sf
M}_{e_j}$. We then reconstruct the sequence $F_j$ (out of the indices $e_j$) within ZFC, where the
rest of the argument happens.

Within ZFC, let ${\cal P}^0$ be given, ordered by the indices $p$ of the ${\sf P}_p$. (See
Definitions \ref {X} and \ref {fel"}.) 
\begin {itemize}
\item {\it Step $0$}\,: We obtain a $f^{\phi}_P$ such that $F_0\neq f^{\phi}_P$. The idea
is to reshuffle ${\cal P}^0$ to suit our purposes. First, generate the list of theorems of ZFC up to
$${\rm ZFC}\vdash\forall x\in\omega\,\exists z\in\omega\,T(e_0,x,z)\,\wedge\,\forall
y\in\omega\,(f(y)=\{e_0\}(y)).$$

Within ZFC, $F_0$ is computed by ${\sf M}_{e_0}$. Get ${\sf M}_{e_0}(0)=y_0'$. Start from ${\cal
P}^0$. Call $f^{\phi}_P(0)$ (computed out of ${\cal P}^0$) the ``provisional'' $f^{\phi}_P(0)$. Now
for the provisional value, either $f^{\phi}_P(0)$ doesn't exist, or $f^{\phi}_P(0)=y_0$. 
\begin {itemize}
\item If the provisional $f^{\phi}_P(0)$ doesn't exist, then $f^{\phi}_P$ is a partial function, and
the proof stops here. Then the provisional value is the definitive one. 
\item If, for the provisional value, $f^{\phi}_P(0)=y_0$, we will ensure that $y_0\neq y'_0$. If
required, we will change things (see below) so that for the definitive value $y_0$ we ensure that
$y_0\neq y'_0$. 
\end {itemize}

There are again two possibilities here:
\begin {itemize}
\item If ${\sf V}(\langle y'_0,{\sf P}^0_0(y'_0)\rangle)=1$, proceed to the next step, and put ${\sf
P}^{\phi}_0={\sf P}^0_0$, the rest being left as in ${\cal P}^0$ until the next step is reached. 

In that case go to an $m=m_1=k_0+1>0$ (see below) and execute the next step.  

\item If ${\sf V}(\langle y'_0,{\sf P}^0_0(y'_0)\rangle)=0$, let ${\sf P}^0_{k_0}$ be the first
${\cal G}$--bounded machine in ${\cal P}^0$ such that $${\sf V}(\langle y'_0,{\sf
P}^0_{k_0}(y'_0)\rangle)=1.$$ (In order to make that test compute ${\sf V}(\langle y'_0,{\sf
P}^0_{j}(y'_0)\rangle)$, $j=0,1,2,\ldots$, up to the first index $j=k_0$ such that ${\sf V}\ldots=1$.
Recall that we deal with total machines and that there are infinitely many machines so that $y'_0$
is acceptable; pick a machine like the one last described in Proposition \ref {o}.) 

If that machine accepts all instances, then the function is partial and the proof stops here. If
not, it will reject some instance $y$, and since $y'$ is accepted, $y'\neq y$. 

Then change places between that machine ${\sf P}^0_{k_0}$, and ${\sf P}^0_0$, so that we obtain an
initial segment of ${\cal P}^{\phi}$, ${\cal P}^{\phi}_0$ given by ${\sf P}^0_{k_0}, {\sf P}^0_1,
{\sf P}^0_2, \ldots, {\sf P}^0_{k_0 - 1}, {\sf P}_0$. (That sequence is relabeled:
$${\sf P}^{\phi}_0={\sf P}^0_{k_0}, {\sf P}^{\phi}_1={\sf P}^0_1, {\sf P}^{\phi}_2={\sf P}^0_2,
\ldots, {\sf P}^0_{k_0 - 1}, {\sf P}^{\phi}_{k_0}={\sf P}^0_0$$ and gives the initial segment ${\cal
P}^{\phi}_0$ of ${\cal P}^{\phi}$.) Clearly here $F_0(0)\neq f^{\phi}_P(0)$, since
$f^{\phi}_P$ either diverges at $0$ or differs there from $F_0$. The definitive value of
$f^{\phi}_P(0)$ is also determined here out of the rearranged sequence of the ${\sf P}$s.) 
\end {itemize}
Leave step $0$. 
\item {\it Step $1$}\,: Out of the listing of theorems of ZFC obtain
$${\rm ZFC}\vdash\forall x\in\omega\,\exists z\in\omega\,T(e_1,x,z)\,\wedge\,\forall
y\in\omega\,(f(y)=\{e_1\}(y)).$$

We now obtain $F_1\neq f^{\phi}_P$. Situation is the same as in the first step, with $1$ substituted
throughout for $0$ in the lower indices. For the nontrivial case in Step $)$ we get:    
${\cal P}^{\phi}_0\cup\{{\sf P}^0_{k_1}, {\sf P}^0_{k_0 +2},\ldots, {\sf P}^0_{k_1 - 1}, {\sf
P}^0_{k_0 + 1}\}$ (with a slight abuse of notation), which is the extension ${\cal P}^{\phi}_1$ of
${\cal P}^{\phi}_0$ achieved at the present step. 
\item \ldots
\item  {\it Step $n$}\,: Obtain
$${\rm ZFC}\vdash\forall x\in\omega\,\exists z\in\omega\,T(e_n,x,z)\,\wedge\,\forall
y\in\omega\,(f(y)=\{e_n\}(y)).$$

We get $F_n\neq f^{\phi}_P$. \ldots . Same procedure. \Vox 
\end {itemize}

\begin {cla}
${\cal P}^{\phi}$, the reordered sequence of ${\cal G}$--bounded machines that results from the 
extension of all initial segments ${\cal P}^{\phi}_m$, is a recursive sequence of ${\cal G}$--bounded
machines. 
\end {cla}

{\it Proof}\,: The map $\phi :{\cal P}^0\rightarrow {\cal P}^0$ is clearly $1$--$1$ and total: let
us be given ${\sf P}_n\in{\cal P}$. To compute $\phi ({\sf P}_n)$, obtain an ordered initial
segment ${\cal P}^{\phi}_j$ as above of cardinality $>n+1$; $\phi ({\sf P}_n)$ is the machine at
position $n$ in that segment ${\cal P}^{\phi}_m$. The converse operation is immediate, given ${\cal
P}^{\phi}$, out of the definition of ${\cal P}^0$. \Vox 

\begin {cla}
For each $i$, ${\rm ZFC}\vdash f^{\phi}_P\neq F_i$. \Vox
\end {cla}

\begin {cla}\label {xot}
For each $i$, ${\rm ZFC}\vdash\exists x\,[\neg P^{\phi}(x,F_i(x))]$. \Vox
\end {cla}

\begin {cor}\label {ek3}
${\rm ZFC}$ is $\Sigma _1$--sound (see Definition \ref {sig}) if and only if ${\rm ZFC}\not\vdash
f^{\phi}_P$ {\rm is total}. 
\end {cor}

{\it Proof}\,: We have shown that a proof of ``$f^{\phi}_P$ {\rm is total}'' cannot appear in the
listing of all proofs of ZFC. \Vox

\ 

>From the preceding claims and from Proposition \ref {ekiv} and Corollaries \ref {ek2} and \ref
{ek3}: 

\begin {cor}\label {finis}
If {\rm ZFC} is $\Sigma _1$--sound then ${\rm ZFC}${\rm\ plus }{\rm\ [The partial
recursive function }$f^0_P${\rm\ isn't total]} is consistent. \Vox
\end {cor}

\markboth {da Costa, Doria}{Total recursive function, I}
\section {Application: results about the $NP$ class}
\markboth {da Costa, Doria}{Total recursive function, I}

We consider here the $NP$ class of problems. 

\subsection*{The satisfiability problem}
\addcontentsline {toc}{subsection}{The satisfiability problem}

\begin {rem}\rm 
The motivation here comes from the satisfiability problem for Boolean expressions in conjunctive
normal form (cnf). 

Suppose that we write a predicate $A^{\phi}(m,x)$ which means ``polynomial machine ${\sf
P}^{\phi}_m$ accepts $x$,'' that is, ${\sf P}^{\phi}_m$ inputs a binary string $x$ which
polynomially codes a Boolean expression in cnf and outputs another binary string 
$s$ which satisfies that same instance $x$:
\begin {itemize}
\item We can in fact polynomially encode each Boolean expression in cnf as a binary string; that
coding may be constructed as an onto map ${\rm Sat}\rightarrow\omega$, where ${\rm Sat}$ is the set
of all such Boolean expressions (tautologies and totally false expressions excluded).  
\item If $x\in{\rm Sat}$, adequately encoded, and if ${\sf P}_n$ is the $n$--th polynomial
machine in the listing given above, let us write $s={\sf P}_n(x)$. 
\item ${\sf V}$ is the ``verifying machine'' that polynomially checks whether $s$ satisfies $x$ or
not. That is the meaning of ${\sf V}(x,s)=1$ ($s$ satisfies $x$) or ${\sf V}(x,s)=0$ ($s$ doesn't
satisfy $x$). 
\end {itemize}
The conditions we impose on ${\sf V}$ (see Definition \ref {v}) arise out of that motivation.
\Vox 
\end {rem}

\subsection*{Polynomial machines}
\addcontentsline {toc}{subsection}{Polynomial machines}

\begin {rem}\rm \label {pol}
Let the set ${\cal G}$ (Definition \ref {G}) be the set of polynomials:
\begin {enumerate}
\item $p_0(x)=x^3 + 3$.
\item $p_n(x)=x^{n+3}+(n+3)$. \Vox
\end {enumerate}
\end {rem}

\begin {prop}
The set of ${\cal G}$--bounded Turing machines whose bounds are given by Remark \ref {pol} is
(essentially) the set of all polynomial machines. 
\end {prop}

{\it Proof}\,: Every polynomial machine can be represented as a pair $\langle {\sf
M}_n,[p_k]\rangle$. And every Turing machine coupled to a polynomial clock such as $\langle {\sf
M}_n,[p_k]\rangle$ is polynomial. \Vox

\

\noindent (See \cite {Bak}.) Throughout this Section, ${\cal P}^0$ is a set of expressions
that represent polynomial Turing machines. Let ${\sf V}(x,s)$ be the (fixed) polynomial machine that
checks whether $s$ solves instance $x$ (Definition \ref {v}). If ${\sf V}(x,s)=1$, we say that $s$
satisfies $x$.

\subsection*{The $NP$ class of problems}
\addcontentsline {toc}{subsection}{The $NP$ class of problems}

Next definition characterizes the $NP$ class \cite {Bak,Stock}:

\begin {tef}\label {np}
Let $x, y$ be binary words (identified to the corresponding numbers), let $p$ be a polynomial, and
let $R(x,y)$ be a 2--place recursive polynomial predicate. Then:
\begin {enumerate}
\item $x\in A_{p,R}\leftrightarrow \exists y\,[|y|\leq p(|x|)\wedge\, R(x,y)]$. 
\item $NP=\{A_{p,R}:p$ {\rm is a polynomial and $R$ is a 2--place recursive polynomial predicate}\}.
\Vox
\end {enumerate}
\end {tef} 
Then, after Definition \ref {np}: 

\begin {tef}
$x\in {\rm SAT}\leftrightarrow\exists s\,[|s|\leq p_*(|x|)\wedge {\sf P}(x)=s\wedge\,{\sf
V}(\langle x,s\rangle)=1]$, where $p_*$ is a fixed polynomial, and ${\sf P}$ is a polynomial
machine. \Vox
\end {tef}

\begin {rem}\rm
The condition $p_*(|x|)\geq |s|$ directly follows from the satisfiability problem. \Vox
\end {rem}

\begin {tef}
For ${\sf P}^{\phi}_m\in{\cal P}^{\phi}$, 
$$A^{\phi}(m,x)\leftrightarrow_{\rm Def}[{\sf V}(\langle x,{\sf
P}^{\phi}_m(x)\rangle)=1]\leftrightarrow\exists s\,[{\sf P}^{\phi}_m(x)=s\wedge\,{\sf
V}(\langle x,s\rangle)=1].\,\mbox {\Vox}$$
\end {tef}

\begin {tef}
{\bf (Formalization of $P<NP$ for {\rm SAT}.)} The $P<NP$ conjecture (for {\rm SAT}) in {\rm ZFC} is:
$$\forall m\in\omega\,\exists x\in\omega\,\neg A^0(m,x).\,\mbox{\Vox}$$ 
\end {tef}

\begin {prop}\label {totus}
If $f^0_{\neg A}(m)=\mu _x\neg A^0(m,x)$, then 
$${\rm ZFC}\vdash (\forall m\in\omega\,\exists x\in\omega\,\neg A^0(m,x))\leftrightarrow [f^0_{\neg
A}\mbox{\rm\ is total}].\,\mbox{\Vox}$$
\end {prop}

\subsection*{If {\rm ZFC} is $\Sigma _1$--unsound then ${\rm ZFC}\vdash P<NP$}
\addcontentsline {toc}{subsection}{If {\rm ZFC} is $\Sigma _1$--unsound then ${\rm ZFC}\vdash P<NP$} 

\begin {lem}\label {Kr}
Let $R$ be a recursively axiomatizable simply consistent extension of ${\rm PA}$ which is not $\Sigma
_1$--sound. Then every partial recursive function is equal to a function which is provably total
from $R$. That is, for each $e$, there is an $a$ such that $\varphi _e =\varphi _a$, and $R\vdash
[\mbox {\rm $\varphi _a$ is total}]$. 
\end {lem}

{\it Proof}\,: To construct $a$, fix a primitive recursive predicate $Q(x)$ such that
$R\vdash\exists x\,Q(x)$, but such that $Q(n)$ is false for each natural number $n$. (Possible,
since $R$ isn't $\Sigma _1$--sound.) In Kleene Normal Form \cite {IM}, we have $\varphi _e (x) =
U(\mu _y T(e,x,y))$, where $U$, $T$ are as in the reference; recall that they are primitive
recursive. Then define $\varphi _a(x) = U[\mu _y (T(e,x,y)\vee Q(y))]$. $\varphi _a=\varphi _e$,
since $Q(y)$ is always false, but $R\vdash\forall x\,\exists y\,[T(e,x,y)\vee\,Q(y)]$, so $R\vdash 
[\mbox {\rm $\varphi _a$ is total}]$. \Vox

\begin {rem}\rm
In order to obtain one such $R$, for example pick up PA and if $P$ is primitive recursive and
$\forall x\,\neg Q(x)$ is true while independent of PA, take $R={\rm PA}+\exists x\,Q(x)$. \Vox
\end {rem}

\begin {cor}
If {\rm ZFC} isn't $\Sigma _1$--sound, then ${\rm ZFC}\vdash P<NP$. 
\end {cor}

{\it Proof}\,: From Proposition \ref {totus} and Lemma \ref {Kr}. \Vox 

\subsection*{If {\rm ZFC} is $\Sigma _1$--sound then ${\rm ZFC}\not\vdash P<NP$}
\addcontentsline {toc}{subsection}{If {\rm ZFC} is $\Sigma _1$--sound then ${\rm ZFC}\not\vdash
P<NP$}

\begin {prop}
If ${\rm ZFC}$ is $\Sigma_1$--sound, then:
\begin {enumerate}
\item $f^0_{\neg A}$ isn't {\rm ZFC}--provably total. 
\item ${\rm ZFC}\not\vdash P<NP$.
\item ${\rm ZFC}$ is consistent if and only if ${\rm ZFC} + (P=NP)$ is consistent.
\end {enumerate}
\end {prop}

{\it Proof}\,: From Proposition \ref {totus} and Corollary \ref {finis}. \Vox

\begin {cor}\label {FIN} If {\rm ZFC} is consistent, then {\rm ZFC} is $\Sigma _1$--sound if and only
if ${\rm ZFC}\not\vdash P<NP$. \Vox
\end {cor}

\begin {rem}\rm
We've just added Corollary \ref {FIN} as a special assertion in order to emphasize the depth of the
$P?NP$ question, which leads to such a beautiful interplay between formal systems and the
surrounding metamathematics. The previous results are certainly valid for any $\Sigma _1$--sound
fragment of ZFC which contains PA. \Vox
\end {rem}

\subsection*{On the relativized result ${\rm ZFC}\vdash P^A<NP^A$}
\addcontentsline {toc}{subsection}{On the relativized result ${\rm ZFC}\vdash P^A<NP^A$}

\begin {rem}\rm
We comment here on the well--known results in \cite {Bak}. Some preliminary remarks must be made
before we get into details: 
\begin {itemize}
\item We first notice that the Turing machines considered in the present paper paper are of a more
restricted kind than those in the reference, as (in our case) the only special state is $s_0$ (see
Remark \ref {TM}). Actually our special machine ${\sf V}$ stands for their ``accepting states,''
with several restrictions added (see Definition \ref {v}). 

The restrictions we impose on ${\sf V}$ are of course modelled on the way one mechanically verifies
the validity of a choice of truth values for Boolean expressions in cnf, and on their possible
choices. 
\item We also recall that, for oracle machines where the oracle is realized as an extra tape in a
multitape Turing machine with the oracle set written on it, if $P=NP$ then $P^A=NP^A$, for any 
oracle $A$ (\cite {Bak}, Remark in p.\ 437). 
\item Yet in \cite {Bak} one supposes that the oracle is consulted, and gives its answer, in a
single step. 
\end {itemize}

But the point is, in which way (if any) does our construction in Proposition \ref {sel} conflict
with, say, the one in the same Theorem 3? 
\begin {itemize}
\item First of all, in their proof of Theorem 3 they use the oracle in an essential way, that is,
all machines must {\it nontrivially} refer to the oracle during their computations. That is to say,
we exclude situations like the machine with the oracle tape which is never accessed by its program,
or the machine that proceeds in the same direction never caring whether it does get a ``yes'' or
``no'' from the oracle. 

So, from start we notice that the freedom we enjoyed in moving machines to and fro in the proof of
Proposition \ref {sel} may not be available in the (very restrictive) relativized setting, as we
proceed with the constructions in the reference. 
\item But let us consider the proof of Theorem 3 in the reference \cite {Bak}: in Theorem 3 one aims
at the construction of a language $L(B)$ which will be rejected by every polynomial oracle machine
${\sf P}_i^B$ which nontrivially consults with oracle $B$. That is to say, at least one instance
$x\in L(B)$ isn't accepted by each ${\sf P}_i^B$. 
\item If $B$ is the oracle, then $L(B)=\{y:|y|=|x|\,\wedge x\in B\}$. 
\item The proof in the reference proceeds by a stepwise construction of $B$; if step $i$ precedes
step $j$, then $B_i\subseteq B_j$, and $B=\bigcup _i B_i$. Moreover, $B$ can be made recursive. 
\item Once one has $L(B)$, we can obtain a provably total recursive function $h_B(i)=\min
(y:|y|=|x_i|\,\wedge x_i\in B)$. 
\item It is clear that $h_B(i)\neq f^B_{\neg A}(i)$. 
\end {itemize}

The essential point is: it is not at all clear whether we would be able to make the permutations
which are essential to the proof of Proposition \ref {sel} in the present restricted, relativized
setting. \Vox
\end {rem}

\subsection*{On the diagonalization procedure used here}
\addcontentsline {toc}{subsection}{On the diagonalization procedure used here}

\begin {rem}\rm
We must also add a few comments on the diagonalization used here. Diagonal constructions have been
used out of a `lower' class $X$ to obtain objects in a `higher' class $Y\supset X$. More precisely,
out of some listing of classes of $NP$--languages which are known to be in $P$, we try to obtain
via a diagonal construction a language which isn't in $P$. 

Our procedure in this paper elaborates on a well--known construction (see \cite {Ku}, p.\ 54, 
Ex.\ 5). We temporarily forget about $P$ and $NP$ and just consider the relation between $f^0_P$
and the listing of all ZFC--provably total unary recursive functions. We also try to impose some
properties on that $f^0_P$, such that it can be proved to be Turing--computable, and one can
prove that there is a G\"odel number for the corresponding machine. 

The diagonal construction is then used to show that such a function differs from each one of the
ZFC--provable t.r.u.\ functions at least once. To put it in another way, instead of using a single
sequence of functions to diagonalize out of it, we couple two sequences, the fundamental one---the
sequence of the $F_i$---and the subordinate sequence, the ${\sf P}_j$. \Vox 
\end {rem}

\subsection*{$G?NG$}
\addcontentsline {toc}{subsection}{$G?NG$}

\begin {rem}\rm
M.\ Benda has pointed out to the authors that their original construction leads to a whole hierarchy
of problems of the form $G?NG$, which is why we have decided to frame our results in that wider
setting from start \cite {Ben}. \Vox
\end {rem}

\markboth {da Costa, Doria}{Total recursive function, I}
\section {Acknowledgments}
\markboth {da Costa, Doria}{Total recursive function, I}

Notice that both Definition \ref {sig} and Lemma \ref {Kr} are crucial to this work. On the proof
of Proposition \ref {sel} see \cite {Ku}. 

Portions of this work were written while the second author was on leave at the Graduate Studies
Program in Communications at the Federal University at Piau\'\i\ and later at FACOM, Federal
University at Juiz de Fora (Brazil). Their hospitality is thankfully acknowledged. Formatting of
this paper is due to {\it Project Griffo} at the Rio de Janeiro Federal University; thanks are due
to E.\ Carneiro Le\~ao, Muniz Sodr\'e, R.\ Paiva, J.\ Argolo and P.\ Pires. 

Internet facilities for FAD were provided by the M.\ Kritz Research Program at the LNCC--CNPq; its
support and help is gratefully acknowledged. 

Finally both authors wish to thank CNPq, CAPES, and FAPESP, Philosophy Section, for support of the
present work. 

\bibliographystyle {plain}

\end{document}